\documentclass[pdflatex,sn-mathphys,iicol]{sn-jnl}

\usepackage{graphicx}%
\usepackage{multirow}%
\usepackage{amsmath,amssymb,amsfonts}%
\usepackage{amsthm}%
\usepackage{mathrsfs}%
\usepackage[title]{appendix}%
\usepackage{xcolor}%
\usepackage{textcomp}%
\usepackage{manyfoot}%
\usepackage{booktabs}%
\usepackage{algorithm}%
\usepackage{algorithmicx}%
\usepackage{algpseudocode}%
\usepackage{listings}%
\usepackage{array}

\theoremstyle{thmstyleone}%
\theoremstyle{thmstyletwo}%

\theoremstyle{thmstylethree}%

\begin{document}

\title[Space Domain based Ecological Cooperative and Adaptive Cruise Control on Rolling Terrain]{Space Domain based Ecological Cooperative and Adaptive Cruise Control on Rolling Terrain}

\author[1]{\fnm{Mingyue} \sur{Lei}}\email{mingyue\_l@tongji.edu.cn}

\author*[1,2]{\fnm{Haoran} \sur{Wang}}\email{wang\_haoran@tongji.edu.cn}

\author[3]{\fnm{Lu} \sur{Xiong}}\email{xiong\_lu@tongji.edu.cn}

\author[4]{\fnm{Jaehyun (Jason)} \sur{So}}\email{jso@ajou.ac.kr}

\author[5]{\fnm{Ashish} \sur{Dhamaniya}}\email{adhamaniya@gmail.com}

\author[1]{\fnm{Jia} \sur{Hu}}\email{hujia@tongji.edu.cn}

\affil[1]{\orgdiv{Key Laboratory of Road and Traffic Engineering of the Ministry of Education}, \orgname{Tongji University},  \city{Shanghai}, \country{China}}

\affil[2]{\orgdiv{State Key Laboratory of Advanced Design and Manufacturing for Vehicle Body}, \orgname{Hunan University}, \city{Changsha}, \country{China}}

\affil[3]{\orgdiv{School of Automotive Studies}, \orgname{Tongji University}, \city{Shanghai}, \country{China}}

\affil[4]{\orgdiv{Department of Transportation Systems Engineering}, \orgname{Ajou University}, \city{Gyeonggi-do}, \country{Republic of Korea}}

\affil[5]{\orgdiv{Department of Civil Engineering}, \orgname{SV National Institute of Technology}, \city{Gujarat}, \country{India}}


\abstract{Cooperative and Adaptive Cruise Control (CACC) is widely focused to enhance driving fuel-efficiency by maintaining a close following gap. The ecology of CACC could be further enhanced by adapting to the rolling terrain. However, current studies cannot ensure both planning optimality and computational efficiency. Firstly, current studies are mostly formulated on the conventional time domain. These time domain based methods cannot ensure planning optimality for space-varying road slopes. Secondly, fuel consumption models are non-linear and hard to solve efficiently. Hence, this paper proposes a space domain based Ecological-CACC (Eco-CACC) controller. It is formulated into a nonlinear optimal control problem with the objective of optimizing global fuel consumptions. Furthermore, a differential dynamic programming-based solving method is developed to ensure real-time computational efficiency. Simulation results have shown that the proposed Eco-CACC controller can improve average fuel saving by 37.67\% at collector road and about 17.30\% at major arterial. String stability of the proposed method has been theoretically proven and experimentally validated.}

\keywords{Ecological driving, Cooperative and Adaptive Cruise Control, Rolling terrain, Spatial domain, Nonlinear optimal control, String stability}



\maketitle

\textbf{\fontsize{12pt}{14pt}\selectfont Abbreviations}

\noindent
CACC \ \ Cooperative and Adaptive Cruise Control

\noindent
Eco-CACC \ \ Ecological-Cooperative and Adaptive Cruise Control

\noindent
ACC \ \ Adaptive Cruise Control

\noindent
MOE \ \ Measurements of Effectiveness

\section{Introduction}
%
%
%
%

Cooperative and Adaptive Cruise Control (CACC) technology emerges as an advancement over Adaptive Cruise Control (ACC) technology \cite{turri2016cooperative} \cite{sakhdari2017distributed} \cite{zhang2020cooperative} \cite{hu2024vehicles} \cite{kazemi2018learning} \cite{tan2023observer} \cite{xiang2024v2x} \cite{meng2024toward} \cite{wang2024towards}. It enables a platoon to maintain a reduced inter-vehicular distance without vehicle speed fluctuations \cite{hu2024safety} \cite{wang2022make}. This capability contributes to diminishing the aerodynamic resistance for the platoon, which has the potential to reduce energy consumptions and emissions. Building upon this, Ecological CACC (Eco-CACC) is introduced with the objectives of enhancing fuel efficiency through the utilization of CACC technology.

However, existing Eco-CACC researches are still with limitations.

Firstly, most of existing Eco-CACC researches lack adaptability in driving on rolling terrain, which is common in road transportation \cite{zhong2023vehicle}. These researches do not take road slope into consideration, so it turns out that adopting the same control policy on different slopes \cite{rezaee2023leaderless} \cite{wang2017developing} \cite{ma2020cooperative} \cite{han2013decentralized} \cite{wang2018review} \cite{zhang2022human}. However, these policies result in wasted power and increased fuel consumption. It is due to that vehicles on different slopes experience different frictional forces and horizontal components of gravity. To follow the constant control policy, the engine of the vehicle is inevitably engaged in compensating for resistance on uphills and idling on downhills. Excessive fuel is consumed without contributing to productive work. Hence, vehicles should be provided with dynamic power, that is related to the road slopes at their positions. Some researches adopt this strategy and introduce road slope into the control system’s dynamics as an external disturbance \cite{li2023energy} \cite{turri2018fuel} \cite{zhai2018ecological}. However, this design lacks planning optimality. It only considers the impact of road slope on vehicle dynamics, without accounting for its significant effects on fuel efficiency. Therefore, it is necessary to develop a controller that ensures planning optimality for space-varying road slopes.

Secondly, the fuel-saving benefits for the entire platoon still have room for improvement. For most of the existing Eco-CACC researches \cite{duan2022study} \cite{lin2019adaptive} \cite{luu2021research}, only the platoon leader's ecology efficiency is optimized. The followers just track the leader's speed as the reference. This design is uncapable to optimally save the followers' fuel consumption, as each vehicle within the platoon has its individual vehicle dynamics and road conditions it encounters. Hence, in order to achieve greater fuel efficiency, it is necessary to develop a centralized system which could optimize all the platoon vehicles' fuel consumption simultaneously.

Thirdly, the solutions in existing Eco-CACC researches face challenges in ensuring both optimality and computational efficiency. The problem of energy optimization for multiple vehicles tends to be modelled as a non-convex and non-linear formulation. Many researches utilize heuristic algorithms to resolve it \cite{zhai2018ecological} \cite{ma2020cooperative} \cite{liu2015v2x}. Heuristic algorithms cannot always guarantee optimal solutions, and the speed of convergence is unstable. There are also some researches dividing the formulation into sub-problems, and utilizing layered optimization methods to derive a solution \cite{murgovski2016cooperative}. This strategy also sacrifices precision for computation efficiency, impeding the performance of Eco-CACC applications. Dong et al. \cite{dong2023cooperative} propose a valuable work on energy-optimal control. This study is one of the few that considers spatio-temporal constraints into cooperative eco-driving. However, this method has to modify the position constraint before each computing, since it is modeled in the conventional temporal domain. Hence, planning infeasibility and optimality may be concerned. Tian et al. \cite{tian2021trajectory} first adopt the arrival time at the sampling spatial position as a decision variable. This study provides an effective approach to conveniently address the spatially dependent constraints. However, the adoption of arrival time as the decision variable leads to strong nonlinearity issues. To handle this challenge, Tian et al. utilizes a relaxed linearization method to transform a nonlinear expression of acceleration into a linear approximation expression. Although easy in computation, this linearization method greatly reduces the solution domain, thereby reducing planning optimality. Furthermore, the approximated acceleration may be not feasible for vehicle local control to fulfill accurately. Hence, there is a significant demand for a solution method that ensures both optimality and computational efficiency.

To address the above deficiencies, this paper proposes an advanced Eco-CACC system that is fine-tuned for the following capabilities:

\begin{itemize}

\item \textbf{Ensuring planning optimality across rolling terrains}: By modeling in the space domain, the proposed controller is capable of accurately capturing the space-varying road slope. Road slope does not have to be updated after trajectory planning in the temporal domain. In this space domain based planning, road slopes could considered in optimization as fixed values at each position. Hence, planning optimality could be ensured.
\item \textbf{Enhanced fuel efficiency}: The proposed controller is capable of globally optimizing all vehicles’ fuel consumptions within the platoon. Individual’s sacrifice is allowed to realize a systematic fuel efficiency enhancement.
\item \textbf{Real-time computational efficiency}: This research develops a differential dynamic programming-based method for problem solving. This algorithm could rapidly provide the optimal solution to non-linear optimal control problems.
\item \textbf{String stability ensured}: The proposed Eco-CACC controller has been mathematically proven and experimentally validated with string stability.

\end{itemize}

The structure of this paper is as follows: Section \uppercase\expandafter{\romannumeral2} provides detailed formulations about the proposed system. Section \uppercase\expandafter{\romannumeral3} reveals the solution to the Eco-CACC controller in the system. Section \uppercase\expandafter{\romannumeral4} provides the string stability analysis of the controller. Section \uppercase\expandafter{\romannumeral5} shows the experiment design and evaluation results. Section \uppercase\expandafter{\romannumeral6} demonstrates discussions. Section \uppercase\expandafter{\romannumeral7} provides conclusions.

\section{PROBLEM FORMULATION}

\subsection{Overall Architecture}
\begin{figure*}[!t]
  \centering
  \includegraphics[scale=0.6]{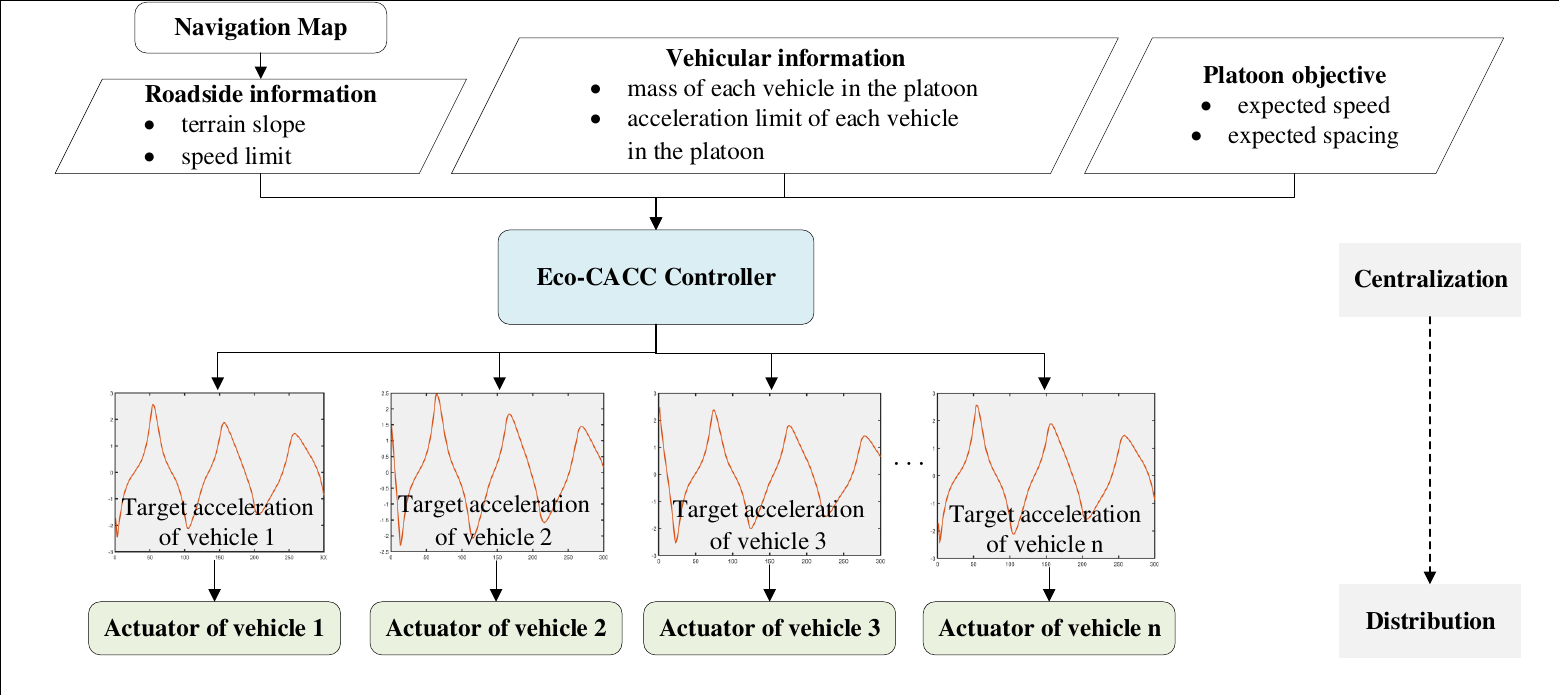}
  \caption{Architecture of the proposed centralization-distribution Eco-CACC system}
  \label{Figure1}
\end{figure*}

The architecture of the proposed Eco-CACC system is depicted in Figure \ref{Figure1}. It adopts a bifurcated structure consisting of an upper-centralized planning level and a lower-distributed actuation level. This paper focuses on the upper-centralized planning. The architecture is demonstrated as follows:

The upper-centralized planning is achieved by the proposed Eco-CACC controller. It integrates both roadside and vehicular information to formulate optimal acceleration commands for each vehicle within the platoon. The roadside information is obtained from the navigation map of the automated vehicle. Parameters such as expected speed and inter-vehicle time spacing are pre-configured within this central controller to guide the platooning behavior.

The lower-distributed actuating is achieved by the individual vehicle's actuator. These actuators are assumed to be already equipped on the automated vehicle. They are capable of real-time executing commands from the proposed Eco-CACC controller.

\subsection{Space Domain based Eco-CACC Controller}
The proposed Eco-CACC controller formulation is introduced.

\subsubsection{State explanation}
\

\textit{\textbf{Definition 1 (Slowness)}}: Slowness is the time a vehicle takes to travel through a unit distance:
\begin{equation}
    \pi \stackrel{\text { def }}{=} \frac{d t}{d s}=\frac{1}{v}
    \label{Equation1}
\end{equation}

The system state vector $\boldsymbol{X}$ is as follows:
\begin{equation}
    \begin{array}{l}
    \left.\boldsymbol{X}_{k}=[t_{1, k}-t_{2, k}-h, \pi_{1, k}-\pi_{2, k},\cdots,\right.\\
    \left.t_{1, k}-t_{i, k}-(i-1) h, \pi_{1, k}-\pi_{i, k}, \cdots,\right.\\
    \left.t_{1, k}-t_{N, k}-(N-1) h, \pi_{1, k}-\pi_{N, k}]^{T},\right.\\
    i \in[2, N], k \in[0, K]
    \label{Equation2}
    \end{array}
\end{equation}
\begin{equation}
    \pi_{i, k}=\frac{1}{v_{i, k}}, i \in[1, N], k \in[0, K]
    \label{Equation3}
\end{equation}    
where $t_{i, k}$ is the $i^{t h}$ vehicle's arriving time at step $k$, $\pi_{i, k}$ is the $i^{t h}$ vehicle's slowness at step $k$, $v_{i, k}$ is the $i^{t h}$ vehicle's speed at step $k$, $N$ is the number of vehicles in the platoon, $h$ is the expected time headway between adjacent vehicles in the platoon, $K$ is the control horizon.

The control vector $\boldsymbol{U}$ is:
\begin{equation}
    \begin{array}{l}
    \left.\boldsymbol{U}_{k}=\left[a_{1, k}, \cdots, a_{i, k}, \cdots, a_{N, k}\right]^{T},\right.\\
    i \in[1, N], k \in[0, K-1]
    \label{Equation4}
    \end{array}
\end{equation}
where $a_{i, k}$ is the $i^{t h}$ vehicle's acceleration at step $k$.

\subsubsection{System dynamics}
\

The dynamics of vehicles is formulated based on a nonlinear bicycle model \cite{rajamani2011vehicle} \cite{hu2024mirroring} \cite{hu2024eco}. The dynamics is described as follows:
\begin{equation}
    \begin{array}{c}
    \left.\boldsymbol{X}_{k+1}=\left(\boldsymbol{A} \cdot \Delta \mathrm{s}+\boldsymbol{I}_{2(N-1) \times 2(N-1)}\right) \boldsymbol{X}_{k}\right.\\
    +\left(\boldsymbol{B}_{k} \cdot \Delta \mathrm{s}\right) \boldsymbol{U}_{k}, k \in[0, K-1]
    \label{Equation5}
    \end{array}
\end{equation}
with
\begin{equation}
    \boldsymbol{A}=\left[\begin{array}{cccc}
    \mathbb{A} & \mathbf{0} & \cdots & \mathbf{0} \\
    \mathbf{0} & \mathbb{A} & \cdots & \mathbf{0} \\
    \vdots & \vdots & \ddots & \vdots \\
    \mathbf{0} & \mathbf{0} & \cdots & \mathbb{A}
    \end{array}\right]_{2(N-1) \times 2(N-1)}
    \label{Equation6}
\end{equation}
\begin{equation}
    \mathbb{A}=\left[\begin{array}{ll}
    0 & 1 \\
    0 & 0
    \end{array}\right]
    \label{Equation7}
\end{equation}
\begin{equation}
    \begin{array}{c}
    \left.\boldsymbol{B}_{k}=\left[\begin{array}{cccc}
    \mathbb{B}_{k}^{1} \mathbb{B}_{2, k}^{2} & \cdots & \mathbf{0} \\
    \vdots & \vdots & \ddots & \vdots \\
    \mathbb{B}_{k}^{1} & \mathbf{0} & \cdots & \mathbb{B}_{N, k}^{2}
    \end{array}\right]_{2(N-1) \times N} \quad,\right.\\
    k \in[0, K-1]
    \label{Equation8}
    \end{array}
\end{equation}
\begin{equation}
    \mathbb{B}_{k}^{1}=\left[\begin{array}{c}
    0 \\
    -\left(\pi_{1, k}\right)^{3}
    \end{array}\right], k \in[0, K-1]
    \label{Equation9}
\end{equation}
\begin{equation}
    \mathbb{B}_{i, k}^{2}=\left[\begin{array}{c}
    0 \\
    \left(\pi_{i, k}\right)^{3}
    \end{array}\right], i \in[2, N], k \in[0, K-1]
    \label{Equation10}
\end{equation}
where $\boldsymbol{A}$ is the state coefficient matrix, $\mathbb{A}$ is the unit block of the diagonal coefficient matrix $\boldsymbol{A}$, $\Delta s$ is the space increment in each step, $\boldsymbol{B}_{k}$ is the control coefficient matrix at step $k$, $\mathbb{B}_{i, k}^{1}$ and $\mathbb{B}_{i, k}^{2}$ are the unit blocks of the diagonal coefficient matrix $\boldsymbol{B}_{k}$.

\subsubsection{Cost function}
\

\begin{equation}
    J=\sum_{i=0}^{K-1} L\left(\boldsymbol{X}_{i}, \boldsymbol{U}_{i}\right)+l^{f}\left(\boldsymbol{X}_{K}\right)
    \label{Equation11}
\end{equation}
with
\begin{equation}
    \begin{array}{c}
    L\left(\boldsymbol{X}_{k}, \boldsymbol{U}_{k}\right)
    =\sum_{i=2}^{N} q_{1}\left(t_{1, k}-t_{i, k}-(i-1) h\right)^{2} \\
    +\sum_{i=1}^{N} q_{2}(m_{i} a_{i, k} v_{i, k}+m_{i} g \sin (\theta_{k}) v_{i, k}\\
    +\mu m_{i} g \cos \left(\theta_{k}\right) v_{i, k}+\xi v_{i, k}{ }^{3}) \\
    +\sum_{i=1}^{N} r_{1} a_{i, k}{ }^{2}, k \in[0, K-1] \\
    \end{array}
    \label{Equation12}
\end{equation}
\begin{equation}
    \begin{array}{c}
    l^{f}\left(\boldsymbol{X}_{K}\right)=\sum_{i=1}^{N} q_{3}\left(t_{i, K}-\frac{K \cdot \Delta \mathrm{s}}{v^{d}}\right)^{2}
    \end{array}
    \label{Equation13}
\end{equation}
where $L(\boldsymbol{X}, \boldsymbol{U})$ is the running cost function, $l^{f}(\boldsymbol{X})$ is the terminal cost function, $q_1$, $q_2$, $q_3$ and $r_1$ are non-negative weighting parameters, $m_i$ is the $i^{t h}$ vehicle's mass, $g$ is gravity, $\theta_{k}$ is the terrain slope at step $k$, which is constant relative to longitudinal position (control step). $\mu$ is a constant that describes rolling resistance, $\xi$ is a constant that describes drag, $v^d$ is the target average speed. $\sum_{i=2}^{N} q_{1}\left(t_{1, k}-t_{i, k}-(i-1) h\right)^{2}$ is a CACC cost. The CACC cost is designed to maintain a desired time gap between adjacent vehicles in the platoon. $\sum_{i=1}^{N} q_{2}(m_{i} a_{i, k} v_{i, k}+m_{i} g \sin (\theta_{k}) v_{i, k}+\mu m_{i} g \cos (\theta_{k}) v_{i, k}+\xi v_{i, k}{ }^{3})$ is an ecology cost. The ecology cost is designed to reduce total fuel consumption of the platoon. The terminal cost is designed to guarantee the mobility of the platoon.

\subsubsection{Boundary conditions}
\

The initial system state is a boundary condition, which is obtained via V2I communication:
\begin{equation}
    \begin{array}{l}
    \left.\boldsymbol{X}_{0}=[t_{1,0}-t_{2,0}-h, \pi_{1,0}-\pi_{2,0}, \cdots,\right. \\
    \left.t_{1,0}-t_{i, 0}-(i-1) h, \pi_{1,0}-\pi_{i, 0}, \cdots,\right. \\
    \left.t_{1,0}-t_{N, 0}-(N-1) h, \pi_{1,0}-\pi_{N, 0}\right]^{T}, i \in[2, N]
    \end{array}
    \label{Equation14}
\end{equation}
where $t_{i,0}$ is the moment that the $i^{t h}$ vehicle in the platoon reaches the starting position, $\pi_{i,0}$ is the slowness of the $i^{t h}$ vehicle in the platoon at the starting position.

\subsubsection{Constraints}
\

(a) Speed limit

Taking the speed limit into account, vehicle speed should be constrained within the following range:
\begin{equation}
    0 \leq v_{i, k} \leq v_{max}, i \in[1, N], k \in[0, K]
    \label{Equation15}
\end{equation}
where $v_{max}$ is the speed limit.

(b) Acceleration limit

Each vehicle's acceleration should be limited considering the individual vehicle's capability:
\begin{equation}
    a_{i,min} \leq a_{i, k} \leq a_{i,max}, i \in[1, N], k \in[0, K-1]
    \label{Equation16}
\end{equation}
where $a_{i,min}$ is the $i^{t h}$ vehicle's minimum acceleration, and $a_{i,max}$ is the $i^{t h}$ vehicle's maximum acceleration.

\section{PROBLEM SOLVING}

\begin{figure*}[thpb]
  \centering
  \includegraphics[scale=0.13]{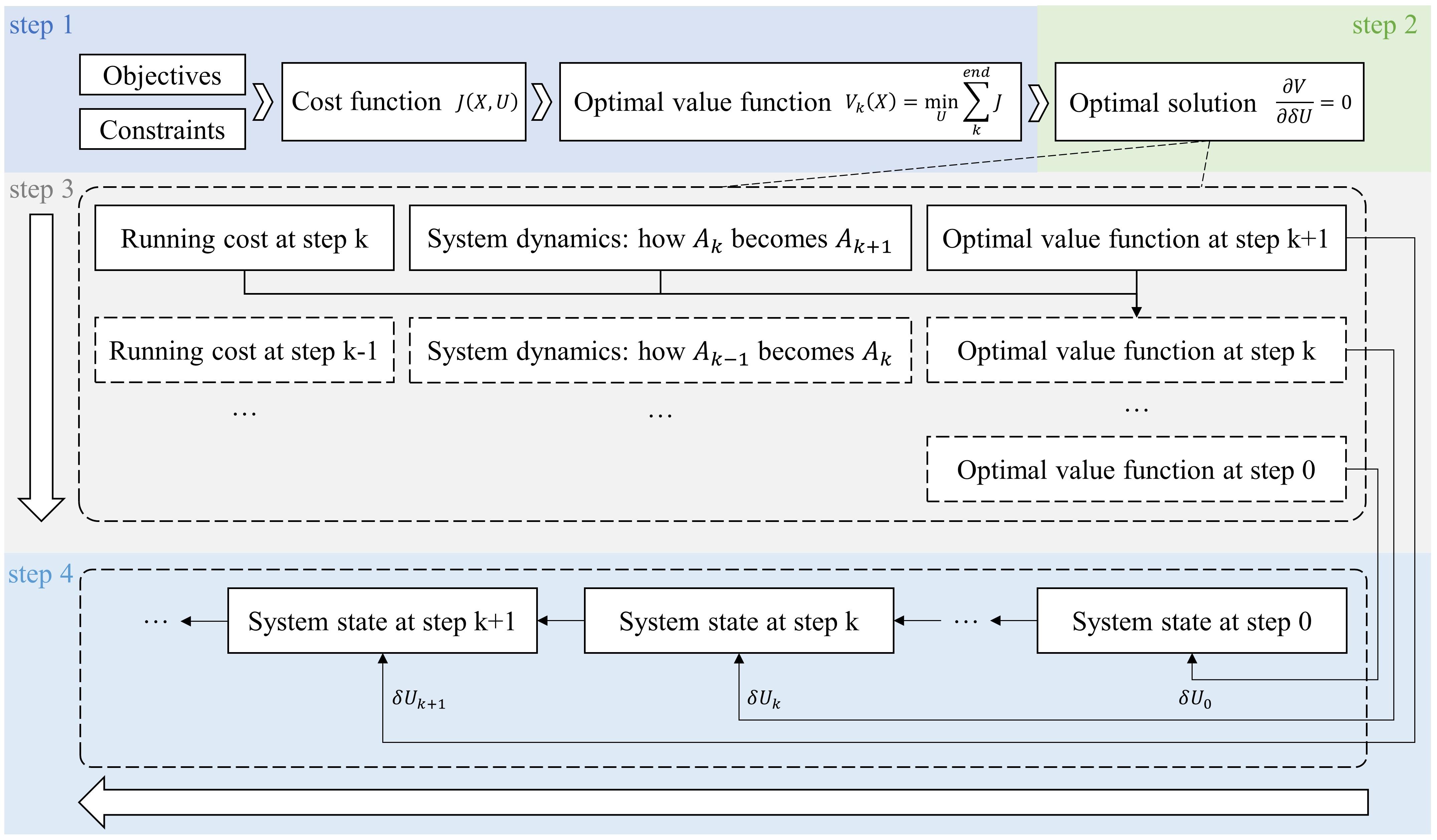}
  \caption{Framework of solution}
  \label{Figure2}
\end{figure*}

The proposed controller is formulated as a non-linear optimal control problem. To ensure computational efficiency, a differential dynamic programming-based method is developed in this section. The detailed process is as follows and the framework of solution is illustrated in Figure \ref{Figure2}.

\subsection{Step 1: Initialization of the Optimal Value Function}

In an optimal control problem, the optimality of future control actions is independent of historical control or state trajectories. Consequently, the optimal value (cost-to-go) function is constructed to reflect the minimum achievable cost from a given state, encapsulating the essence of optimality in this context.

To incorporate constraints within the optimization framework, the Augmented Lagrangian Method is employed. This approach serves to reformulate the cost function in equation (\ref{Equation12}), enabling a more comprehensive solution to the control problem.
\begin{equation}
    J^{P}=\sum_{i=0}^{K-1} L^{P}\left(\boldsymbol{X}_{i}, \boldsymbol{U}_{i}\right)+l^{f}\left(\boldsymbol{X}_{K}\right)
    \label{Equation17}
\end{equation}
with
\begin{equation}
    \begin{array}{c}
    L^{P}\left(\boldsymbol{X}_{k}, \boldsymbol{U}_{k}\right)
    =L\left(\boldsymbol{X}_{k}, \boldsymbol{U}_{k}\right)\\
    +\frac{1}{2} \sum_{i=1}^{\varepsilon} C_{i}\left(\boldsymbol{X}_{k}, \boldsymbol{U}_{k}\right)^{T} \rho_{i} C_{i}\left(\boldsymbol{X}_{k}, \boldsymbol{U}_{k}\right) \\
    +\sum_{i=1}^{\varepsilon} \lambda_{i} C_{i}\left(\boldsymbol{X}_{k}, \boldsymbol{U}_{k}\right), k \in[0, K-1]
    \end{array}
    \label{Equation18}
\end{equation}
\begin{equation}
    \begin{array}{c}
    C_{i}\left(\boldsymbol{X}_{k}, \boldsymbol{U}_{k}\right)=e_{i}\left(\boldsymbol{X}_{k}, \boldsymbol{U}_{k}\right)+s_{i},\\
    i \in[1, \varepsilon], k \in[0, K-1]
    \end{array}
    \label{Equation19}
\end{equation}
where $L^{P}(\boldsymbol{X}, \boldsymbol{U})$ is the running cost function containing the penalty of constraint violations, $C_{i}(\boldsymbol{X}, \boldsymbol{U})$ is the $i^{t h}$ equality constraint, $\varepsilon$ is the number of equality constraints. According to equation (\ref{Equation15}) and equation (\ref{Equation16}), for each vehicle in the platoon, there exist four constraints to be set. Hence, $\varepsilon$ equals to $4*N$ in this research. $\rho$ are non-negative penalty parameters used to balance the objective function and the constraint function, $\lambda$ is the Lagrange multiplier vector for equality constraints. The update for $\lambda_{i}$ shall be presented in the equation (\ref{Equation45}). $e_{i}(\boldsymbol{X}, \boldsymbol{U})$ is the $i^{t h}$ inequality constraint, which can be derived from equation (\ref{Equation15}) and equation (\ref{Equation16}). $s_{i}$ is the slacking variable for $e_{i}(\boldsymbol{X}, \boldsymbol{U})$. The update for $s_{i}$ is presented in the equation (\ref{Equation44}).

According to the cost function equation (\ref{Equation17}), the optimal cost-to-go function at step $k$ is defined as follows.
\begin{equation}
    \begin{array}{l}
    \left. V_{k}(\boldsymbol{X})=\min _{\boldsymbol{u}} \sum_{i=k}^{K-1} L^{P}\left(\boldsymbol{X}_{i}, \boldsymbol{U}_{i}\right)+l^{f}\left(\boldsymbol{X}_{K}\right), \right.\\
    k \in[0, K-1]
    \label{Equation20}
    \end{array}
\end{equation}
where $\boldsymbol{U}=\left\{\boldsymbol{U}_{k}, \boldsymbol{U}_{k+1}, \cdots, \boldsymbol{U}_{K-1}\right\}$ is a sequence of control.

Using Bellman's principle of optimality \cite{bellman1966dynamic} \cite{nocedal1999numerical}, the optimal value function can be expressed in a recursive formulation.
\begin{equation}
    \begin{array}{c}
    V_{k}(\boldsymbol{X})=\min\limits_{\mathbf{\boldsymbol{U}_{k}}} L^{P}\left(\boldsymbol{X}_{k}, \boldsymbol{U}_{k}\right)+V_{k+1}\left(\boldsymbol{f}\left(\boldsymbol{X}_{k}, \boldsymbol{U}_{k}\right)\right),\\
    k \in[0, K-1]
    \end{array}
    \label{Equation21}
\end{equation}
where $\boldsymbol{f}$ is the dynamics function that governs the state transition given the current state and control. The formulation of $\boldsymbol{f}$ can be derived from equation (\ref{Equation5}) to equation (\ref{Equation9}).

The boundary condition is:
\begin{equation}
    V_{K}(\boldsymbol{X})=l^{f}\left(\boldsymbol{X}_{K}\right)
    \label{Equation22}
\end{equation}

Considering the minimization problem in equation (\ref{Equation21}) is non-linear, local approximation to the optimal value function is introduced to settle the computational difficulty caused by nonlinearity.

$Q_{k}(\boldsymbol{\delta}_{\boldsymbol{X}},\boldsymbol{\delta}_{\boldsymbol{U}})$ is the local change in equation (\ref{Equation21}) under perturbation $\left(\boldsymbol{\delta}_{\boldsymbol{X}}, \boldsymbol{\delta}_{\boldsymbol{U}}\right)$, which is defined as:
\begin{equation}
    \begin{array}{c}
    Q_{k}\left(\boldsymbol{\delta}_{\boldsymbol{X}}, \boldsymbol{\delta}_{\boldsymbol{U}}\right) \\
    =L^{P}\left(\boldsymbol{X}_{k}+\boldsymbol{\delta}_{\boldsymbol{X}}, \boldsymbol{U}_{k}+\boldsymbol{\delta}_{\boldsymbol{U}}\right) \\
    +V_{k+1}\left(\boldsymbol{f}\left(\boldsymbol{X}_{k}+\boldsymbol{\delta}_{\boldsymbol{X}}, \boldsymbol{U}_{k}+\boldsymbol{\delta}_{\boldsymbol{U}}\right)\right), k \in[0, K-1]
    \end{array}
    \label{Equation23}
\end{equation}

In order to conduct recursive iteration in the following step, $Q_{k}\left(\boldsymbol{\sigma}_{\boldsymbol{X}}, \boldsymbol{\sigma}_{\boldsymbol{U}}\right)$ is expanded using second-order Taylor about $(\mathbf{0}, \mathbf{0})$:
\begin{equation}
    \begin{array}{c}
    Q_{k}\left(\boldsymbol{\delta}_{\boldsymbol{X}}, \boldsymbol{\delta}_{\boldsymbol{U}}\right) \\
    \approx \frac{1}{2}\left[\begin{array}{c}
    \mathbf{1} \\
    \boldsymbol{\delta}_{\boldsymbol{X}} \\
    \boldsymbol{\delta}_{\boldsymbol{U}}
    \end{array}\right]^{T}\left[\begin{array}{ccc}
    \mathbf{0} & \boldsymbol{Q}_{\boldsymbol{X}, k}^{T} & \boldsymbol{Q}_{\boldsymbol{U}, k}^{T} \\
    \boldsymbol{Q}_{\boldsymbol{X}, k} & \boldsymbol{Q}_{\boldsymbol{X} \boldsymbol{X}, k} & \boldsymbol{Q}_{\boldsymbol{X} \boldsymbol{U}, k} \\
    \boldsymbol{Q}_{\boldsymbol{U}, k} & \boldsymbol{Q}_{\boldsymbol{X} \boldsymbol{U}, k}^{T} & \boldsymbol{Q}_{\boldsymbol{U} U, k}
    \end{array}\right]\left[\begin{array}{c}
    \mathbf{1} \\
    \boldsymbol{\delta}_{\boldsymbol{X}} \\
    \boldsymbol{\delta}_{\boldsymbol{U}}
    \end{array}\right],\\
    k \in[0, K-1]
    \end{array}
    \label{Equation24}
\end{equation}
where the block matrices are computed as:
\begin{equation}
    \boldsymbol{Q}_{\boldsymbol{X}, k}=L_{\boldsymbol{X}, k}^{P}+\boldsymbol{f}_{\boldsymbol{X}}^{T} \boldsymbol{b}_{k+1}
    \label{Equation25}
\end{equation}
\begin{equation}
    \boldsymbol{Q}_{\boldsymbol{U}, k}=L_{\boldsymbol{U}, k}^{P}+\boldsymbol{f}_{\boldsymbol{U}}^{T} \boldsymbol{b}_{k+1}
    \label{Equation26}
\end{equation}
\begin{equation}
    \boldsymbol{Q}_{\boldsymbol{X} \boldsymbol{X}, k}=L_{\boldsymbol{X} \boldsymbol{X}, k}^{P}+\boldsymbol{f}_{\boldsymbol{X}}^{T} \boldsymbol{A}_{k+1} \boldsymbol{f}_{\boldsymbol{X}}+\boldsymbol{b}_{k+1}^{T} \boldsymbol{f}_{\boldsymbol{X} \boldsymbol{X}}
    \label{Equation27}
\end{equation}
\begin{equation}
    \boldsymbol{Q}_{\boldsymbol{U} U, k}=L_{\boldsymbol{U} U, k}^{P}+\boldsymbol{f}_{\boldsymbol{U}}^{T} \boldsymbol{A}_{k+1} \boldsymbol{f}_{\boldsymbol{U}}+\boldsymbol{b}_{k+1}^{T} \boldsymbol{f}_{\boldsymbol{U} \boldsymbol{U}}
    \label{Equation28}
\end{equation}
\begin{equation}
    \boldsymbol{Q}_{\boldsymbol{U} X, k}=L_{\boldsymbol{U} X, k}^{P}+\boldsymbol{f}_{\boldsymbol{U}}^{T} \boldsymbol{A}_{k+1} \boldsymbol{f}_{\boldsymbol{X}}+\boldsymbol{b}_{k+1}^{T} \boldsymbol{f}_{\boldsymbol{U} \boldsymbol{X}}
    \label{Equation29}
\end{equation}
\begin{equation}
    L_{\boldsymbol{X}, k}^{P}=L_{\boldsymbol{X}, k}+\sum_{i=1}^{\varepsilon} e_{i}{ }_{\boldsymbol{X}}^{T} \lambda_{i}+\sum_{i=1}^{\varepsilon} e_{i}{ }_{\boldsymbol{X}}^{T} \rho_{i}\left(e_{i}+s_{i}\right)
    \label{Equation30}
\end{equation}
\begin{equation}
    L_{\boldsymbol{U}, k}^{P}=L_{\boldsymbol{U}, k}+\sum_{i=1}^{\varepsilon} e_{i}{ }_{\boldsymbol{U}}^{T} \lambda_{i}+\sum_{i=1}^{\varepsilon} e_{i}{ }_{\boldsymbol{U}}^{T} \rho_{i}\left(e_{i}+s_{i}\right)
    \label{Equation31}
\end{equation}
\begin{equation}
    L_{\boldsymbol{X} \boldsymbol{X}, k}^{P}=L_{\boldsymbol{X} \boldsymbol{X}, k}+\sum_{i=1}^{\varepsilon} e_{i}{ }_{\boldsymbol{X}}^{T} \rho_{i} e_{i \boldsymbol{X}}
    \label{Equation32}
\end{equation}
\begin{equation}
    L_{\boldsymbol{U} U, k}^{P}=L_{\boldsymbol{U} U, k}+\sum_{i=1}^{\varepsilon} e_{i}^{T} \rho_{i} e_{i \boldsymbol{U}}
    \label{Equation33}
\end{equation}
\begin{equation}
    \begin{array}{c}
    L_{\boldsymbol{U} \boldsymbol{X}, k}^{P}
    =L_{\boldsymbol{U} \boldsymbol{X}, k}+\sum_{i=1}^{\varepsilon} \lambda_{i}^{T} e_{i \boldsymbol{U} \boldsymbol{X}}\\
    +\sum_{i=1}^{\varepsilon} e_{i}{ }_{\boldsymbol{U X}}^{T} \rho_{i}\left(e_{i}+s_{i}\right)
    +\sum_{i=1}^{\varepsilon} e_{i}{ }_{\boldsymbol{U}}^{T} \rho_{i} e_{i \boldsymbol{X}}
    \end{array}
    \label{Equation34}
\end{equation}
where $\boldsymbol{b}_{k+1}$ is the gradient of $V_{k+1}(\boldsymbol{X})$, $\boldsymbol{A}_{k+1}$ is the Hessian of $V_{k+1}(\boldsymbol{X})$.

\subsection{Step 2: Control Law Generation}

In order to solve the optimization in the optimal value function, $Q_{k}\left(\boldsymbol{\sigma}_{\boldsymbol{X}}, \boldsymbol{\sigma}_{\boldsymbol{U}}\right)$ in the equation (\ref{Equation24}) is minimized with respect to $\boldsymbol{\delta}_{U}$. A linear feedback control law can be computed:
\begin{equation}
    \boldsymbol{\sigma}_{\boldsymbol{U}}=\boldsymbol{h}_{k} \boldsymbol{\sigma}_{\boldsymbol{X}}+\boldsymbol{j}_{k}, k \in[0, K-1]
    \label{Equation35}
\end{equation}
with
\begin{equation}
    \boldsymbol{h}_{k}=-\boldsymbol{Q}_{\boldsymbol{U} U, k}{ }^{-1} \boldsymbol{Q}_{\boldsymbol{U} \boldsymbol{X}, k}^{T}
    \label{Equation36}
\end{equation}
\begin{equation}
    \boldsymbol{j}_{k}=-\boldsymbol{Q}_{\boldsymbol{U} U, k}{ }^{-1} \boldsymbol{Q}_{\boldsymbol{U}, k}^{T}
    \label{Equation37}
\end{equation}

\subsection{Step 3: Iteration of the Optimal Value Function}

In order to calculate the optimal control law evaluated at $\boldsymbol{X}_{k}$ recursively, the optimal value function evaluated at $\boldsymbol{X}_{k}$ needs to be obtained according to the equation (\ref{Equation21}). After substituting the equation (\ref{Equation35}) into the equation (\ref{Equation24}) and equation (\ref{Equation23}), the updated $V_{k}(\boldsymbol{X})$ is formulated as:
\begin{equation}
    \begin{array}{c}
    V_{k}(\boldsymbol{X})
    =\frac{1}{2}\left(\boldsymbol{X}-\boldsymbol{X}_{k}\right)^{T} \boldsymbol{A}_{k}\left(\boldsymbol{X}-\boldsymbol{X}_{k}\right)\\
    +\boldsymbol{b}_{k}^{T}\left(\boldsymbol{X}-\boldsymbol{X}_{k}\right)
    +e_{k}, k \in[0, K-1]
    \end{array}
    \label{Equation38}
\end{equation}
with
\begin{equation}
    \begin{array}{c}
    \left. \boldsymbol{A}_{k}=\boldsymbol{Q}_{\boldsymbol{X}
    \boldsymbol{X}, k}+\boldsymbol{h}_{k}^{T} \boldsymbol{Q}_{\boldsymbol{U} \boldsymbol{U}, k} \boldsymbol{h}_{k}+ \right. \\
    \boldsymbol{Q}_{\boldsymbol{U} \boldsymbol{X}, k}^{T} \boldsymbol{h}_{k}+\boldsymbol{h}_{k}^{T} \boldsymbol{Q}_{\boldsymbol{U} \boldsymbol{X}, k}
    \label{Equation39}
    \end{array}
\end{equation}
\begin{equation}
    \boldsymbol{b}_{k}=\boldsymbol{Q}_{\boldsymbol{X}, k}+\boldsymbol{h}_{k}^{T} \boldsymbol{Q}_{\boldsymbol{U} U, k} \boldsymbol{j}_{k}+\boldsymbol{Q}_{\boldsymbol{U} \boldsymbol{X}, k}^{T} \boldsymbol{j}_{k}+\boldsymbol{h}_{k}^{T} \boldsymbol{Q}_{\boldsymbol{U}, k}
    \label{Equation40}
\end{equation}
\begin{equation}
    c_{k}=\frac{1}{2} \boldsymbol{j}_{k}^{T} \boldsymbol{Q}_{\boldsymbol{U} U, k} \boldsymbol{j}_{k}+\boldsymbol{j}_{k}^{T} \boldsymbol{Q}_{\boldsymbol{U}, k}+\boldsymbol{Q}_{k}(\mathbf{0}, \mathbf{0})
    \label{Equation41}
\end{equation}
where $c_k$ is a constant term, that is irrelevant to the solution. $\boldsymbol{b}_{k}$ and $\boldsymbol{A}_{k}$ evaluated at step $k$ serve as components of $\boldsymbol{b}_{k-1}$ and $\boldsymbol{A}_{k-1}$  at the previous step $k-1$. Starting with $\boldsymbol{A}_{K}=l^{f}_{\boldsymbol{X} \boldsymbol{X}}$ and $\boldsymbol{b}_{K}=l^{f}_{\boldsymbol{X}}$, $\boldsymbol{h}_{k}$, $\boldsymbol{j}_{k}$, $\boldsymbol{A}_{k}$, $\boldsymbol{b}_{k}$ can be recursively solved from step $K$ to step 0.

\subsection{Step 4: Update of the System States}

By leveraging the above optimal control law, the system states (vehicle trajectories) are updated. The update is forward passed starting with $\boldsymbol{X}_{0}^{\text {new }}=\boldsymbol{X}_{0}$.
\begin{equation}
    \boldsymbol{U}_{k}^{\text {new }}=\boldsymbol{U}_{k}+\boldsymbol{h}_{k}\left(\boldsymbol{X}_{k}^{\text {new }}-\boldsymbol{X}_{k}\right)+\boldsymbol{j}_{k}, k \in[0, K-1]
    \label{Equation42}
\end{equation}
\begin{equation}
    \boldsymbol{X}_{k+1}^{n e w}=\boldsymbol{f}\left(\boldsymbol{X}_{k}^{\text {new }}, \boldsymbol{U}_{k}^{n e w}\right), k \in[0, K-1]
    \label{Equation43}
\end{equation}

After updating the system states, $s_i$ and $\lambda_i$ are updated for the next iteration. It is designed for improved approximation of $\boldsymbol{X}_{k}^{*}$.
\begin{equation}
    s_{i}=-\rho_{i}{ }^{-1} \lambda_{i}-e_{i}, i \in[1, \varepsilon]
    \label{Equation44}
\end{equation}
\begin{equation}
    \lambda_{i}=\lambda_{i}+e_{i}+s_{i}, i \in[1, \varepsilon]
    \label{Equation45}
\end{equation}

Repeatedly performing the backward (Step 1 to Step 3) - forward (Step 4) process, the algorithm shall converge within a specified tolerance and the optimal state trajectory sequence $\left[\boldsymbol{X}_{1}^{*}, \cdots, \boldsymbol{X}_{k}^{*}, \cdots, \boldsymbol{X}_{K}^{*}\right], k \in[0, K]$ shall be obtained.

\section{STRING STABILITY ANALYSIS}

This section provides proof of string stability for the proposed platoon controller.
Optimal control law is utilized for stability analysis. Pontryagin’s Minimum Principle (PMP) is introduced to calculate the optimal control law. Necessary conditions for the calculation are as follows:
\begin{equation}
    H(\boldsymbol{X}, \boldsymbol{U}, \lambda)=L(\boldsymbol{X}, \boldsymbol{U})+\lambda^{T} f(\boldsymbol{X}, \boldsymbol{U})
    \label{Equation46}
\end{equation}
\begin{equation}
    H\left(\boldsymbol{X}^{*}, \boldsymbol{U}^{*}, \boldsymbol{\lambda}^{*}\right) \leq H(\boldsymbol{X}, \boldsymbol{U}, \boldsymbol{\lambda})
    \label{Equation47}
\end{equation}
\begin{equation}
    -\frac{d \boldsymbol{\lambda}}{d x}=\frac{\partial H}{\partial \boldsymbol{X}}=\frac{\partial f}{\partial \boldsymbol{X}} \boldsymbol{\lambda}+\frac{\partial L}{\partial \boldsymbol{X}}
    \label{Equation48}
\end{equation}
where $H$ is Hamiltonian, $\boldsymbol{X}$ is the system state vector, $\boldsymbol{U}$ is the control vector, and $\boldsymbol{\lambda}$ is a co-state vector. $L(\boldsymbol{X}, \boldsymbol{U})$ is the cost function that has been introduced in the proposed controller, $f(\boldsymbol{X}, \boldsymbol{U})$ is the function of the system dynamics that has been introduced in the proposed controller, $\boldsymbol{X}^{*}$,$\boldsymbol{U}^{*}$,$\boldsymbol{\lambda}^{*}$ are the optimal solutions.

Applying equation (\ref{Equation12}):
\begin{equation}
    L(\boldsymbol{X}, \boldsymbol{U})=\boldsymbol{X}^{T} \boldsymbol{M} \boldsymbol{X}+\boldsymbol{N} \boldsymbol{X}+\boldsymbol{U}^{T} \boldsymbol{R} \boldsymbol{U}+\boldsymbol{D} \boldsymbol{U}+\boldsymbol{F}
    \label{Equation49}
\end{equation}
with
\begin{equation}
    \boldsymbol{M}=\left[\begin{array}{cccc}
    \mathbb{M} & \mathbf{0} & \cdots & \mathbf{0} \\
    \mathbf{0} & \mathbb{M} & \cdots & \mathbf{0} \\
    \vdots & \vdots & \ddots & \vdots \\
    \mathbf{0} & \mathbf{0} & \cdots & \mathbb{M}
    \end{array}\right]_{2(N-1) \times 2(N-1)}
    \label{Equation50}
\end{equation}
\begin{equation}
    \mathbb{M}=\left[\begin{array}{cc}
    q_{1} & 0 \\
    0 & 0
    \end{array}\right]
    \label{Equation51}
\end{equation}
\begin{equation}
    \boldsymbol{N}=\left[\begin{array}{c}
    0 \\
    (\sin \theta+\mu \cos \theta) m_{2} g v_{1} v_{2} \\
    \ldots \\
    0 \\
    (\sin \theta+\mu \cos \theta) m_{N} g v_{1} v_{N}
    \end{array}\right]_{1 \times 2(N-1)}
    \label{Equation52}
\end{equation}
\begin{equation}
    \boldsymbol{R}=r_{1} \cdot \boldsymbol{I}_{N \times N}
    \label{Equation53}
\end{equation}
\begin{equation}
    \boldsymbol{D}=\left[\begin{array}{lll}
    q_{2} m_{1} v_{1} & \cdots & q_{2} m_{N} v_{N}
    \end{array}\right]_{1 \times N}
    \label{Equation54}
\end{equation}
\begin{equation}
    \boldsymbol{F}=\sum_{i=1}^{N}\left(m_{i} g \sin \theta+\mu m_{i} g \cos \theta\right) \cdot v_{1}
    \label{Equation55}
\end{equation}

Applying equation (\ref{Equation5}) and adding perturbation:
\begin{equation}
    f(\boldsymbol{X}, \boldsymbol{U})=\boldsymbol{A} \boldsymbol{X}+\boldsymbol{B} \boldsymbol{U}+\boldsymbol{C}
    \label{Equation56}
\end{equation}
with
\begin{equation}
    \boldsymbol{C}=\left[\begin{array}{lllll}
    0 & \delta & \ldots & 0 & \delta
    \end{array}\right]_{2(N-1) \times 1}^{T}
    \label{Equation57}
\end{equation}
where $\delta$ is a perturbation on the platoon leader's speed.

As the formulation of the proposed platoon controller is an optimal control problem, equation (\ref{Equation47}) indicates:
\begin{equation}
    \frac{\partial \boldsymbol{H}}{\partial \boldsymbol{U}^{*}}=\mathbf{0}
    \label{Equation58}
\end{equation}

By substituting equation (\ref{Equation46}) (\ref{Equation49}) (\ref{Equation56}) into equation (\ref{Equation48}) (\ref{Equation57}), the optimal control law of the proposed controller is:
\begin{equation}
    \boldsymbol{U}^{*}=-\boldsymbol{R}^{-1}\left(\boldsymbol{D}^{T}+\boldsymbol{B}^{T} \boldsymbol{\lambda}\right)
    \label{Equation59}
\end{equation}
\begin{equation}
    -\frac{d \lambda}{d t}=\boldsymbol{M} \boldsymbol{X}+\boldsymbol{N}^{T}+\boldsymbol{A}^{T} \boldsymbol{\lambda}
    \label{Equation60}
\end{equation}

In order to analyze string stability more effectively, Laplace transform is adopted to transform the above equations into frequency domain.

\textit{\textbf{Definition 2 (Laplace transform)}}: A Time Function $g(x)$ with $x \geq 0$ can be mapped to a complex function in the frequency domain as follows:
\begin{equation}
    G(s)=\mathcal{L}\{g(x)\}=\int_{0}^{\infty} e^{-s x} g(x) d x
    \label{Equation61}
\end{equation}
\begin{equation}
    s \stackrel{\text { def }}{=} \sigma + j \omega
    \label{Equation62}
\end{equation}
where $\sigma$ is the real part, $j$ is the imaginary number, and $\omega$ is the imaginary part.

By applying Definition 2, $\boldsymbol{X}(x)$, $\boldsymbol{U}(x)$, and $\boldsymbol{\lambda}(x)$ are transformed into frequency domain as follows:
\begin{equation}
    \boldsymbol{X}(s) \stackrel{\text { def }}{=} \mathcal{L}\{\boldsymbol{X}(x)\}
    \label{Equation63}
\end{equation}
\begin{equation}
    \boldsymbol{U}(s) \stackrel{\text { def }}{=} \mathcal{L}\{\boldsymbol{U}(x)\}
    \label{Equation64}
\end{equation}
\begin{equation}
    \boldsymbol{r}(s) \stackrel{\text { def }}{=} \mathcal{L}\{\boldsymbol{\lambda}(x)\}
    \label{Equation65}
\end{equation}

By applying Definition 2 and equation (\ref{Equation63}) (\ref{Equation64}) (\ref{Equation65}), the frequency domain formulations of equation (\ref{Equation56}) (\ref{Equation59}) (\ref{Equation60}) are as follows:
\begin{equation}
    \boldsymbol{X}=(s \boldsymbol{I}-\boldsymbol{A})^{-1}(\boldsymbol{B} \boldsymbol{U}+\boldsymbol{C})
    \label{Equation66}
\end{equation}
\begin{equation}
    \boldsymbol{U}=-\boldsymbol{R}^{-1}\left(\boldsymbol{D}^{T}+\boldsymbol{B}^{T} \boldsymbol{r}\right)
    \label{Equation67}
\end{equation}
\begin{equation}
    \boldsymbol{r}=-\left(s \boldsymbol{I}+\boldsymbol{A}^{T}\right)^{-1}\left(\boldsymbol{M} \boldsymbol{X}+\boldsymbol{N}^{T}\right)
    \label{Equation68}
\end{equation}

Substituting equation (\ref{Equation66}) (\ref{Equation68}) into (\ref{Equation67}):
\begin{equation}
    \boldsymbol{U}=(\boldsymbol{P M T B}-\boldsymbol{R})^{-1}\left(\boldsymbol{D}^{T}-\boldsymbol{P} \boldsymbol{N}^{T}-\boldsymbol{P M T C}\right)
    \label{Equation69}
\end{equation}
with
\begin{equation}
    \boldsymbol{P}=\boldsymbol{B}^{T}\left(s \boldsymbol{I}+\boldsymbol{A}^{T}\right)^{-1}
    \label{Equation70}
\end{equation}
\begin{equation}
    \boldsymbol{T}=(s \boldsymbol{I}-\boldsymbol{A})^{-1}
    \label{Equation71}
\end{equation}

\textit{\textbf{Definition 3}}: Considering a platoon with one leader and $N$-1 followers, the platoon is string stable if and only if:
\begin{equation}
    \Gamma_{N, j, j-1} \stackrel{\text { def }}{=} \frac{\left\|\Delta a_{j}(s)\right\|_{2}}{\left\|\Delta a_{j-1}(s)\right\|_{2}} \leq 1, \forall j \in[2, N]
    \label{Equation72}
\end{equation}
where $\Gamma_{N, j, j-1}$ is the acceleration oscillation transfer function between follower $j$-1 and $j$ in a platoon with $N$ vehicles. $\Delta a_{j}(s)$ is the derivative of the $i^{t h}$ vehicle's acceleration with respect to perturbation, with $\Delta a_{j}(s)=\frac{\partial a_{j}(s)}{\partial \delta}$.

Substituting equation (\ref{Equation8}) (\ref{Equation50}) (\ref{Equation52}) (\ref{Equation53}) (\ref{Equation54}) (\ref{Equation57}) (\ref{Equation70}) (\ref{Equation71}) into (\ref{Equation69}), and assuming that: i) $q_1$, $q_2$ and $r_1$ have the same order of magnitude, ii) any $v_{i}(\forall i \in[1, N])$ has the same order of magnitude, iii) any $m_{i}(\forall i \in[1, N])$ has the same order of magnitude, the following equation can be calculated:
\begin{equation}
    \Gamma_{N, j, 1}=\frac{\left\|\Delta a_{j}(s)\right\|_{2}}{\left\|\Delta a_{1}(s)\right\|_{2}}=\left\|\frac{1}{N-1}\right\|_{2}<1, \forall j \in[2, N]
    \label{Equation73}
\end{equation}

The stability criteria in Definition 3 is satisfied. Hence, the proposed Eco-CACC ensures that the platoon remains string stable, regardless of its size. This concludes the proof.

\section{EVALUATION}

The effectiveness of the proposed Eco-CACC system has undergone comprehensive evaluation. The assessments are categorized as follows: (i) an examination of the platoon's fuel efficiency on rolling terrains, (ii) an evaluation of the platoon's stability, and (iii) a measurement of the controller's computational efficiency.

\subsection{Experiment Design}

\subsubsection{Test bed}
\

A MATLAB-based simulation platform serves as the test bed. An 800-meter-long straight road featuring 4 undulating slopes is simulated to mimic hilly terrain.

\subsubsection{Scenario design}
\

\begin{figure}[thpb]
  \centering
  \includegraphics[scale=0.24]{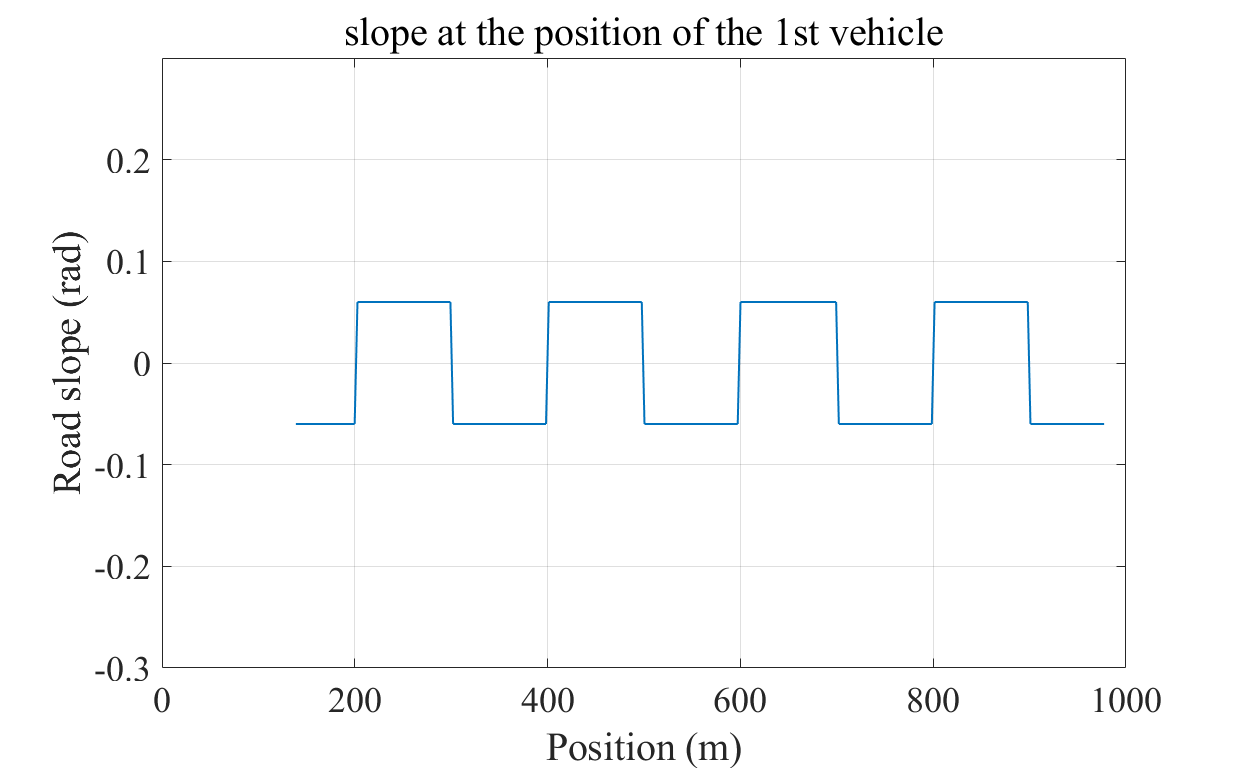}
  \caption{Road slope along driving positions}
  \label{Figure3}
\end{figure}

The scenario involves a platoon of three vehicles navigating this hilly road, with the objective to minimize total fuel consumption while maintaining a specified time gap between vehicles. The slope of the hilly road is designed as shown in Figure \ref{Figure3}.

\subsubsection{Sensitivity analysis}
\

This analysis examines the adaptability of the proposed Eco-CACC system to different road types:
\begin{itemize}
\item {Major arterial}: The target speed is set at 65 mph, with a maximum slope intensity of 6\%.
\item {Collector road}: The target speed is set at 45 mph, with a maximum slope intensity of 15\%.
\end{itemize}

The slope assigned to each road type is in accordance with the guideline from the "Green Book" \cite{hancock2013policy}.

\subsubsection{Controller types}
\

Two distinct controller paradigms are subjected to evaluation:
\begin{itemize}
\item {Traditional CACC controller}: The traditional CACC controller is previously developed by the research team \cite{zhang2022human}. This controller's primary objective is to maintain a minimal and constant inter-vehicular distance. This controller takes vehicle’s actual acceleration as the control variable. Acceleration is executed by local control via transforming into torques as shown in equation (\ref{Equation74}). Hence, it also accounts for the impact of road slopes on the system dynamics. However, it lacks the capability to respond to changing road slopes based on fuel efficiency optimization. The platoon adopting the traditional CACC controller is defined as CACC platoon.
\begin{equation}
    T=\left(a+g \sin (\theta)+\mu g \cos (\theta)+\frac{\xi v^{2}}{m}\right) \cdot m \cdot R
    \label{Equation74}
\end{equation}
where $T$ is the torque, $R$ is the tire radius, $a$, $g$, $\theta$, $\mu$, $\xi$, $v$ and $m$ are defined the same as in Section 2.2.

\item {The proposed Eco-CACC controller}: This proposed controller is designed with the explicit capability to account for variable road slopes, particularly suited for rolling terrains. Furthermore, it incorporates an optimization algorithm that seeks to minimize the energy consumption of each vehicle individually, thereby promoting an enhanced level of global optimality across the platoon. The platoon adopting the proposed Eco-CACC controller is defined as Eco-CACC platoon. 
\end{itemize}

\subsubsection{Parameter settings}
\

The parameters in the proposed Eco-CACC controller are set as Table \ref{Table}.
\begin{table}[ht]
  \centering
    \caption{Parameter Settings}
    \label{Table}
    \fontsize{7pt}{10pt}
    \setlength{\tabcolsep}{0.1pt} 
    \begin{tabular}{>{\centering\arraybackslash}p{0.8cm} >{\centering\arraybackslash}p{4.2cm} >{\centering\arraybackslash}p{1.2cm} >{\centering\arraybackslash}p{1cm}}
        \toprule
        Parameter  & Explanation & Default value & Units\\
        \midrule
        $\Delta s$         & Space increment                            & 0.1                    & meter          \\
        $v_{\max}$         & Speed limit                                & 75                     & mph            \\
        $a_{i, \max }$     & The vehicle's maximum acceleration        & 3                      & $\mathrm{m} / \mathrm{s}^{2}$          \\
        $a_{i, \min }$     & The vehicle's minimum acceleration       & -5                     & $\mathrm{m} / \mathrm{s}^{2}$          \\
        $h$               & Expected time headway                       & 1                      & second         \\
        $q_1$              & Weighting parameter for CACC cost          & 500                    & /              \\
        $q_2$              & Weighting parameter for ecology cost       & 10                     & /              \\
        $q_3$              & Weighting parameter for terminal cost      & 5000                   & /              \\
        $m_i$              & Mass of the vehicle                        & 1400                   & kg             \\
        $g$                & Gravity                                    & 9.8                    & $\mathrm{m} / \mathrm{s}^{2}$          \\
        $\mu$              & Constant describing the rolling resistance & 0.015                  & /              \\
        $\xi$              & Constant describing drag                   & 0.000024               & kg/m           \\
        \bottomrule
    \end{tabular}
\end{table}

\subsubsection{Measurements of effectiveness}
\

Measurements of Effectiveness (MOEs) are categorized as follows:
\begin{itemize}
\item {Platoon's fuel efficiency}: Fuel efficiency is quantified by the accumulated fuel consumption. It is calculated by a VT-micro model \cite{rakha2004development}.
\item {Platoon's stability}: Platoon stability is evaluated by the longitudinal following errors within a platoon. It is calculated by $t_{1, k}-t_{i, k}-(i-1) h$ in equation (\ref{Equation2}).
\item {Computational efficiency}: The computational efficiency is measured by the execution time required for the proposed Eco-CACC controller.
\end{itemize}

\subsection{Results}

The experimental findings validate the capabilities of the proposed Eco-CACC controller, showcasing its ability to: i) improve platoon's fuel efficiency on rolling terrains by 17.30\% $\sim$ 37.67\%, ii) ensure platoon's stability, and iii) maintain an average computation time within 55 milliseconds.

\subsubsection{Platoon's fuel efficiency}
\

\begin{figure}[thpb]
  \centering
  \includegraphics[scale=0.24]{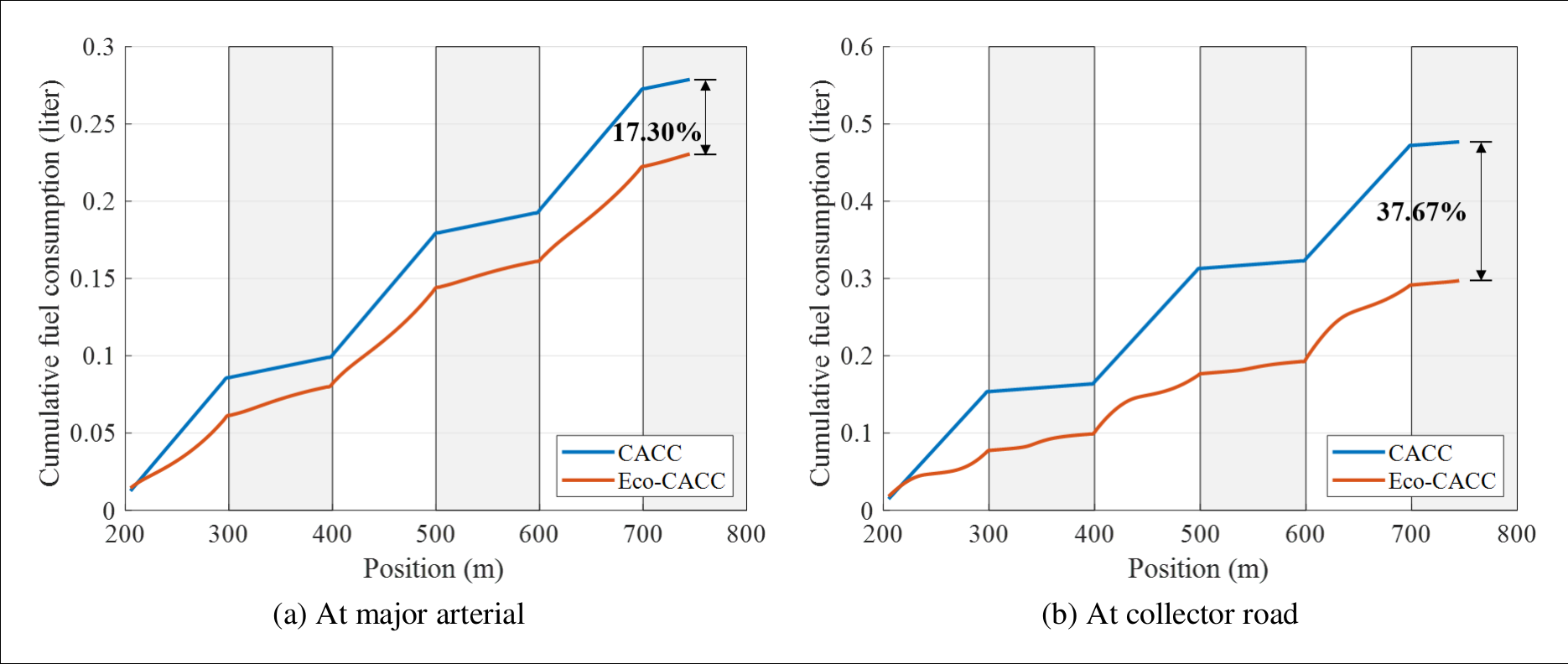}
  \caption{Cumulative fuel consumption of Eco-CACC and CACC}
  \label{Figure4}
\end{figure}

The proposed Eco-CACC controller is capable of enhancing the platoon's fuel efficiency by 17.30\% on major arterials and 37.67\% on collector roads. This superiority is depicted in Figure \ref{Figure4}, illustrating the cumulative fuel consumption of both the Eco-CACC platoon and the baseline CACC platoon. Furthermore, the ecological benefit of the proposed Eco-CACC controller exhibits an expanding trend with increasing travelling distance. This finding reinforces the sustainable advantage of the proposed Eco-CACC system in large-scale applications.

Figure \ref{Figure4} further reveals that the ecological benefit of the proposed Eco-CACC controller is particularly obvious during uphill travel. It is confirmed by the greater disparity between Eco-CACC platoon and CAC platoon during uphill travel (with a white background) compared to downhill travel (with a gray background). This effect is attributed to that Eco-CACC platoon could overcome resistance during uphill by harnessing the kinetic energy accumulated during downhill, without consuming extra fuel energy.

\begin{figure}[thpb]
  \centering
  \includegraphics[scale=0.14]{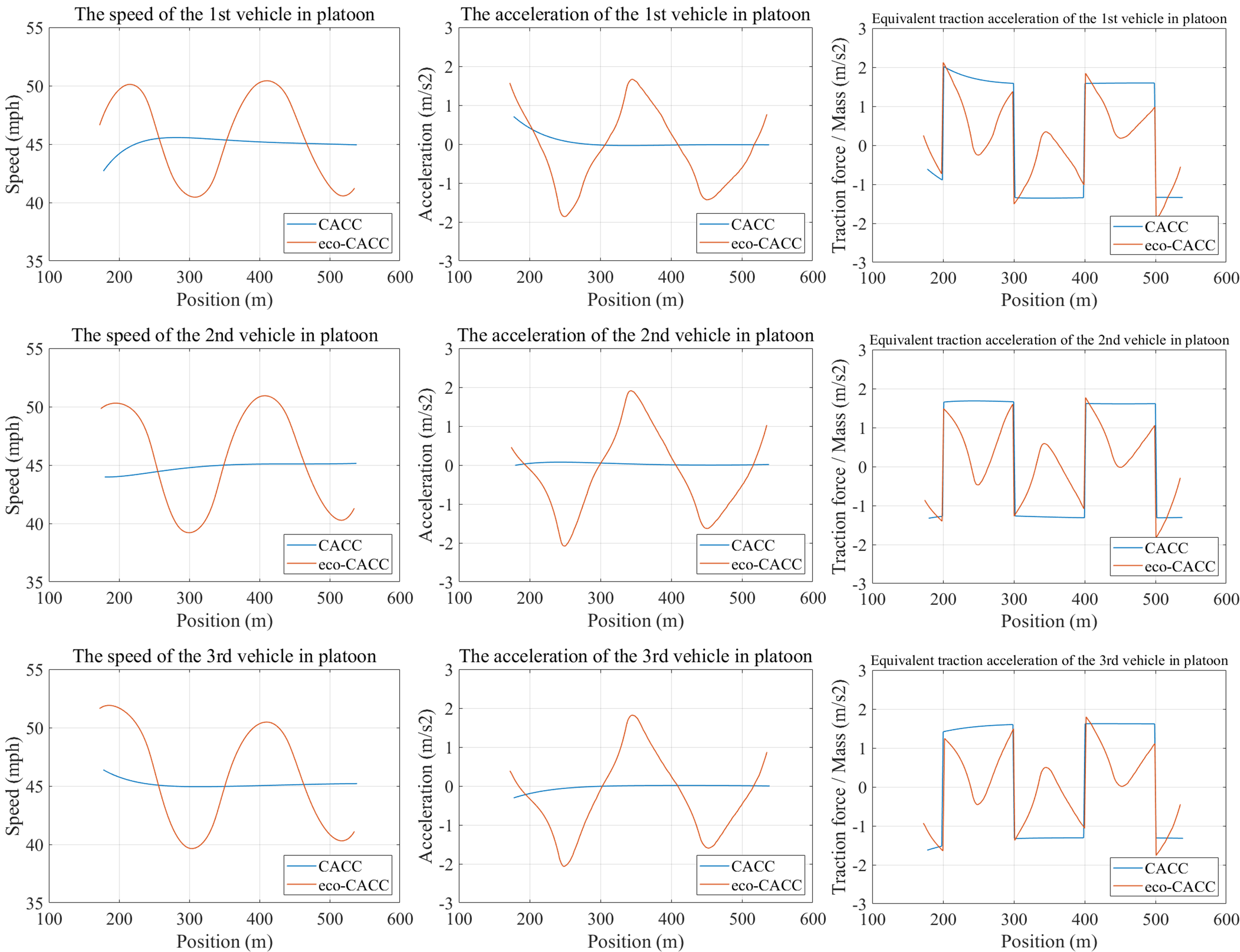}
  \caption{Speeds and accelerations of the vehicles in Eco-CACC platoon and the baseline CACC platoon at collector road}
  \label{Figure5}
\end{figure}

The comparison of vehicles' trajectories between the Eco-CACC platoon and the baseline CACC platoon at collector road is presented, as shown in Figure \ref{Figure5}. By comparing Figure \ref{Figure4} and Figure \ref{Figure5}, it is demonstrated that the proposed Eco-CACC controller's actions adapting to different road slopes contribute to fuel savings. For example, when the road slope is positive (from position 200$m$ to position 300$m$, and from position 400$m$ to position 500$m$), the vehicles in Eco-CACC platoon gradually reduce speed compared to vehicles in the baseline CACC platoon. For a vehicle in Eco-CACC platoon, kinetic energy is converted into gravitational potential energy during the uphill process. The vehicle's speed is allowed to decrease rather than remain constant, consuming less fuel energy to maintain the kinetic energy. However, when the road slope is negative (from position 300$m$ to position 400$m$), the vehicles in Eco-CACC platoon gradually increase speed compared to vehicles in the baseline CACC platoon. For a vehicle in Eco-CACC platoon, gravitational potential energy is converted into kinetic energy during the downhill process. The vehicle's speed is allowed to increase, without consuming much fuel energy to brake the vehicle's motions. The acceleration shown in Figure \ref{Figure5} is the vehicle's actual acceleration. It is an observed vehicle state, influenced by vehicle’s traction force and resistance force on slopes. The equivalent traction acceleration shown in Figure \ref{Figure5} is calculated as the traction force divided by the mass of the vehicle, which compensates for the resistance force in addition to the actual acceleration. Vehicle’s energy consumption is calculated based on the equivalent traction acceleration. As shown in Figure \ref{Figure5}, Eco-CACC platoon exhibits lower amplitude in terms of the equivalent traction acceleration compared to CACC platoon, resulting in reduced fuel consumption.

\subsubsection{Platoon's stability}
\

\begin{figure*}[!t]
    \centering
    \includegraphics[width=\linewidth]{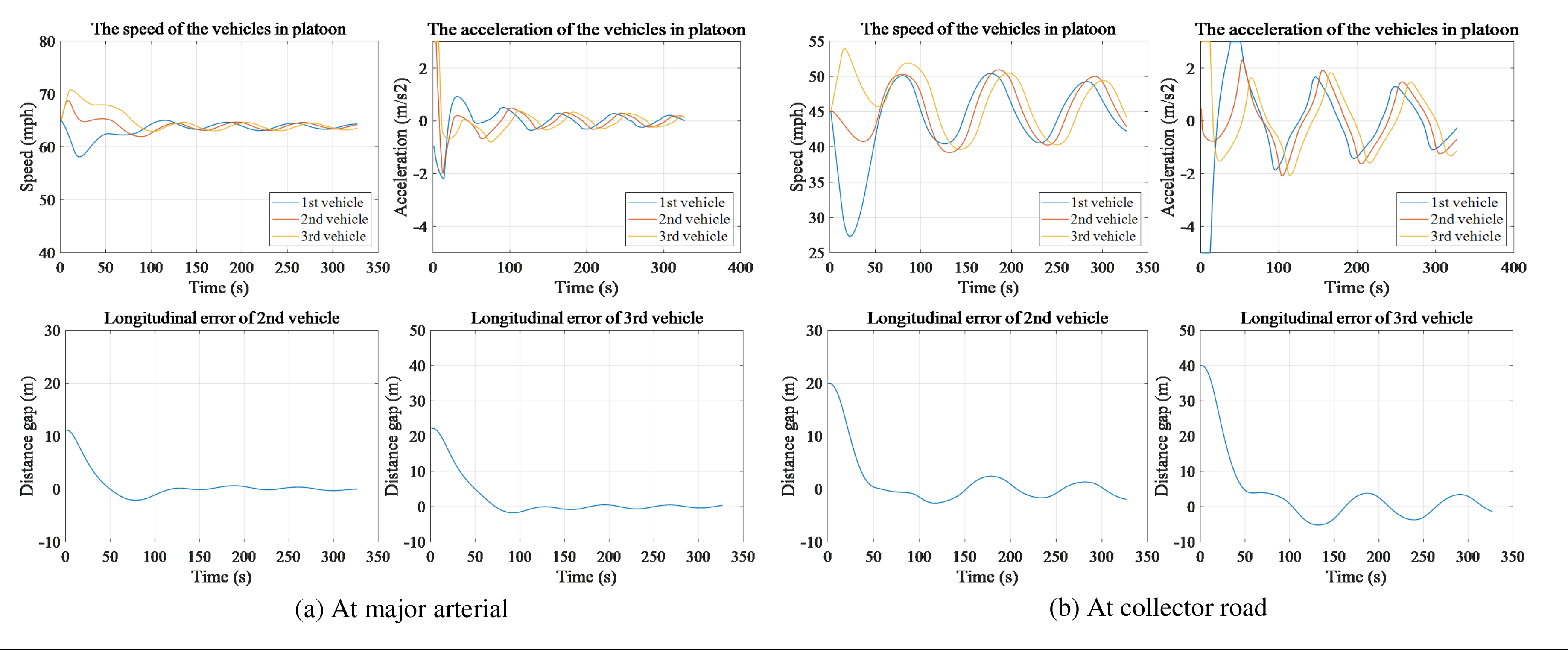}
    \caption{Trajectories of the Eco-CACC platoon (a sample case)}
    \label{Figure6}
\end{figure*}

The proposed Eco-CACC controller is capable of guaranteeing the platoon's stability. It is confirmed by the longitudinal following errors, as shown in Figure \ref{Figure6}. Figure \ref{Figure6} demonstrates a sample case of Eco-CACC platoon on both major arterial and collector road. It highlights that the longitudinal errors of the followers in the platoon diminish along full up-down slopes. Eco-CACC platoon has the ability to maintain a stable spacing.

\subsubsection{Computational efficiency}
\

\begin{figure}[thpb]
  \centering
  \includegraphics[scale=0.44]{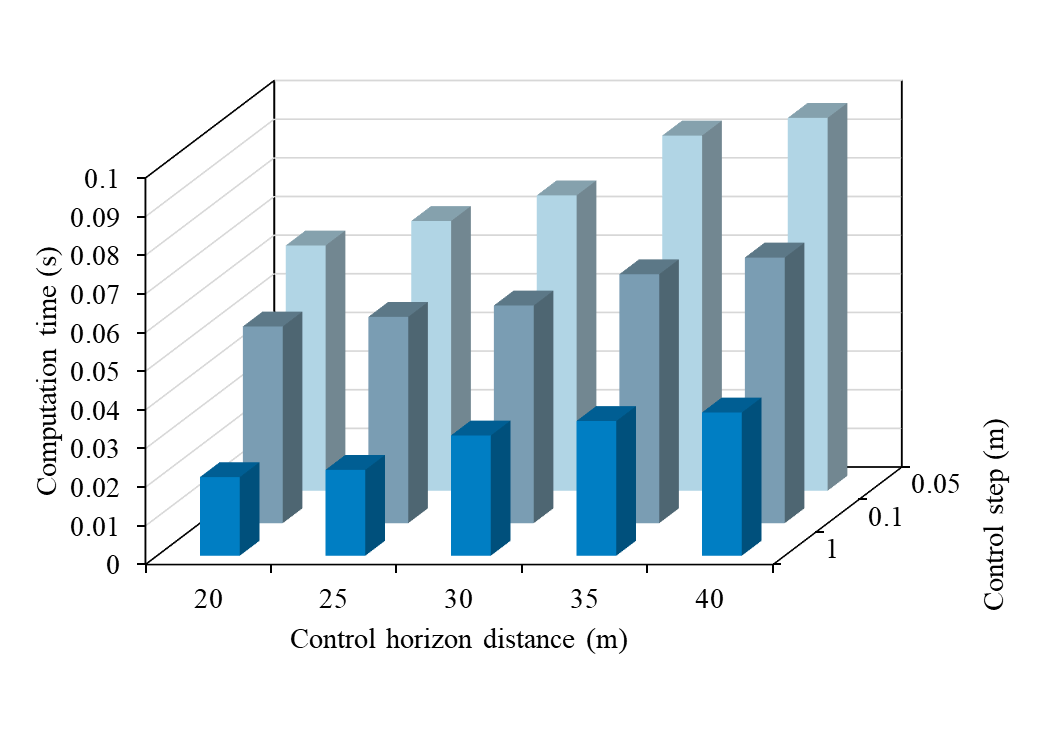}
  \caption{The computation time for an execution of the proposed Eco-CACC controller}
  \label{Figure7}
\end{figure}

Despite its effective performance, the proposed Eco-CACC controller can still guarantee computational efficiency. It is confirmed by the computation time for an execution of the controller, as shown in Figure \ref{Figure7}. It is demonstrated that the computation time is less than 0.096$s$ over various control steps and control horizons. The average computation time is just 0.055$s$, showcasing the online implementation capability of the controller.

Specifically, as the control step ($\Delta s$) decreases from 1$m$ to 0.05$m$, the number of variables in the controller model increases, resulting in longer computation time. However, even though the control step is reduced to 0.05$m$ (a very tiny value that is not typically required in practical applications), the computation time remains under 0.096$s$. In general, the controller with control step of 1$m$ is adopted, and its average computation time is just 0.029$s$.

As the control horizon ($\Delta s \cdot K$) increases from 20$m$ to 40$m$, the total number of variables in the controller model also increases, leading to a slight rise in computation time. However, with a control step of 1$m$, the maximum computation time across different control horizons remains within 0.037$s$. Hence, the proposed method could ensure computational efficiency and has the potential of real-time field implementation.

\section{DISCUSSION}

According to the performances of the proposed Eco-CACC system in the experiments, both mobility and fuel efficiency improvement can be achieved when applying the proposed Eco-CACC system to road transportation. The mobility improvement is born by trajectory optimization coordinating the states of the vehicles in the platoon. The fuel efficiency improvement is born by trajectory optimization coordinating road slopes. Furthermore, when transportation managers give priority to mobility, the weight of cooperative cruising in the objective function of the proposed Eco-CACC controller shall be increased. When transportation managers give priority to fuel efficiency, the weight of reducing fuel consumption in the objective function of the proposed Eco-CACC controller shall be increased.

\section{CONCLUSION}

This paper proposes a nonlinear optimal control based Eco-CACC controller. It bears the following features: i) enhancing performance across rolling terrains by modeling in space domain; ii) enhancing fuel efficiency via globally optimizing all vehicle's fuel consumptions; iii) ensuring computational efficiency by developing a differential dynamic programming-based solving method for the non-linear optimal control problem; iv) ensuring string stability through theoretically proving and experimentally validating. 

The performance of the proposed Eco-CACC controller was evaluated. The influence of road types (different speed limits and slopes) on the performance of the proposed Eco-CACC controller and traditional CACC controller was analyzed. Experiment results showed that the proposed controller is capable of benefiting fuel savings while ensuring the effectiveness of platooning. Detailed analysis on the results further shows that:
\begin{itemize}
\item The proposed Eco-CACC controller significantly improves fuel economy, achieving average fuel savings of approximately 37.67\% on collector road and about 17.30\% on major arterial.
\item The proposed Eco-CACC controller is designed for applications on rolling terrains. It is confirmed by the superior fuel-saving benefits during uphills compared with the traditional CACC controller.
\item Other than fuel efficiency improvement, The proposed Eco-CACC controller is capable of maintaining a stable spacing to ensure the effectiveness of platoon control.
\end{itemize}

However, the proposed controller still needs further consideration, such as involving lateral movement into the problem of ecological platooning. Further work could consider longitudinal and lateral coupled control, for the purpose of further enhancing fuel economy and enabling the platoon navigating through the background traffic.

\

\noindent
\textbf{\fontsize{12pt}{14pt}\selectfont Compliance with Ethical Standards}

\noindent
\textbf{Conflict of interest }On behalf of all the authors, the corresponding author states that there is no conflict of interest.

\bibliographystyle{plain}
\bibliography{sn-bibliography}

\end{document}